\documentclass{article}

\usepackage[main, final]{neurips_2025}

\usepackage{graphicx}
\usepackage{amsmath}
\usepackage{algorithm}
\usepackage{algorithmic}
\usepackage{multirow}
\usepackage{enumitem}

\usepackage[utf8]{inputenc} % allow utf-8 input
\usepackage[T1]{fontenc}    % use 8-bit T1 fonts
\usepackage{hyperref}       % hyperlinks
\usepackage{url}            % simple URL typesetting
\usepackage{booktabs}       % professional-quality tables
\usepackage{amsfonts}       % blackboard math symbols
\usepackage{nicefrac}       % compact symbols for 1/2, etc.
\usepackage{microtype}      % microtypography
\usepackage{xcolor}         % colors

\title{Proxy Target: Bridging the Gap Between Discrete Spiking Neural Networks and Continuous Control}

\author{%
  Zijie Xu \\
  Peking University\\
  Beijing, China 100871 \\
  \texttt{zjxu25@stu.pku.edu.cn} \\
  \And
  Tong Bu \\
  Peking University \\
  Beijing, China 100871\\
  \texttt{putong30@pku.edu.cn} \\
  \AND
  Zecheng Hao \\
  Peking University \\
  Beijing, China 100871\\
  \texttt{haozecheng@pku.edu.cn} \\
  \And
  Jianhao Ding \\
  Peking University \\
  Beijing, China 100871\\
  \texttt{djh01998@alumni.pku.edu.cn} \\
  \And
    Zhaofei Yu\thanks{Corresponding author}\\
  Peking University \\
  Beijing, China 100871\\
  \texttt{yuzf12@pku.edu.cn} \\
}

\begin{document}

\maketitle

\begin{abstract}
  Spiking Neural Networks (SNNs) offer low-latency and energy-efficient decision making on neuromorphic hardware, making them attractive for Reinforcement Learning (RL) in resource-constrained edge devices. However, most RL algorithms for continuous control are designed for Artificial Neural Networks (ANNs), particularly the target network soft update mechanism, which conflicts with the discrete and non-differentiable dynamics of spiking neurons. We show that this mismatch destabilizes SNN training and degrades performance. To bridge the gap between discrete SNNs and continuous-control algorithms, we propose a novel proxy target framework. The proxy network introduces continuous and differentiable dynamics that enable smooth target updates, stabilizing the learning process. Since the proxy operates only during training, the deployed SNN remains fully energy-efficient with no additional inference overhead. Extensive experiments on continuous control benchmarks demonstrate that our framework consistently improves stability and achieves up to $32\%$ higher performance across various spiking neuron models. Notably, to the best of our knowledge, this is the first approach that enables SNNs with simple Leaky Integrate and Fire (LIF) neurons to surpass their ANN counterparts in continuous control. This work highlights the importance of SNN-tailored RL algorithms and paves the way for neuromorphic agents that combine high performance with low power consumption. Code is available at \url{https://github.com/xuzijie32/Proxy-Target}.
\end{abstract}

\section{Introduction}
\begin{figure}[htbp]
  \centering
\includegraphics[width=1\linewidth]{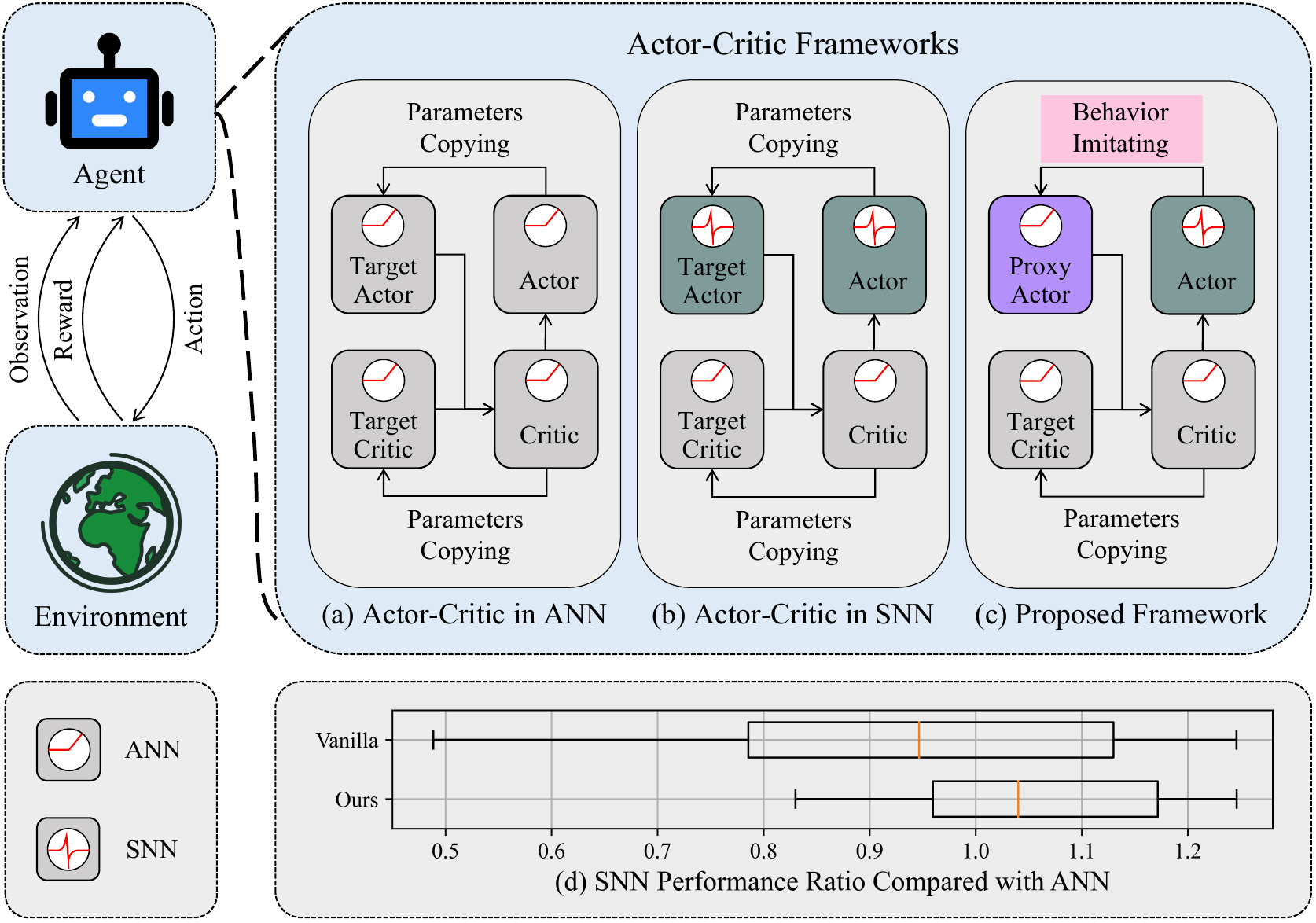}
  \caption{Overview of the training framework and performance comparison. (a)-(c) are different training paradigms. (a) Actor-Critic framework in ANNs, (b) The Actor-Critic framework for SNNs, (c) the proposed proxy target framework for SNNs. (d) Performance ratio of SNNs relative to ANNs across five random seeds and five environments. The middle orange line denotes the median, the box spans from the first to the third quartile, and the whiskers extend to the farthest data within $1.5$ inter-quartile range from the box.}
  \label{fig:big_fig}
\end{figure}

Reinforcement Learning (RL), combined with Artificial Neural Networks (ANNs), has become a cornerstone of modern artificial intelligence, achieving remarkable success in diverse domains such as game playing~\citep{Atari,Go,Mnih2015HumanlevelCT}, autonomous driving~\citep{autodrive1,autodrive2,autodrive3}, and large language model training~\citep{llm1,llm2,llm3}. Among these, continuous control problems have drawn particular attention due to their close alignment with real-world robotic and embodied AI applications~\citep{robot1,robot2,robot3}. However, the high computational cost and power demands of ANN-based RL algorithms limit their deployment on edge devices such as drones, wearables, and IoT sensors~\citep{SNNadvantage,SDDPG,SNNadvantage2}.

Inspired by biological neural systems, Spiking Neural Networks (SNNs) offer sparse, event-driven computation with ultra-low latency and energy consumption on neuromorphic hardware~\citep{debole2019truenorth,davies2018loihi}. These properties make SNNs attractive for RL applications on resource-constrained edge devices~\citep{SNNadvantage2}. Recent works have attempted to integrate SNNs into continuous-control RL algorithms via hybrid frameworks~\citep{SDDPG,popSAN,BPT_SAN,ILC_SAN,MDC_SAN}, where a spiking actor network (SAN) is co-trained with an ANN critic using Spatio-Temporal Backpropagation (STBP)~\citep{STBP,fang2021incorporating}, as illustrated in Fig.~\ref{fig:big_fig}(b). With well-chosen hyperparameters, such frameworks have shown that SNNs can approach or even surpass the performance of ANNs in some tasks.

However, most of these studies simply retrofit SNNs into existing ANN-centric RL frameworks without adapting the algorithms to SNN dynamics. Since ANNs and SNNs exhibit fundamentally different computational characteristics, it remains unclear \textbf{whether RL algorithms designed for continuous, differentiable activations are well-suited for discrete, event-driven networks}. 
% \textbf{Is there a fundamental mismatch between the discrete dynamics of SNNs and the continuous control RL algorithms?}

A key issue arises from the \emph{target network soft update mechanism}, a core component widely used in off-policy RL algorithms to stabilize training by gradually updating target networks \citep{RL1,DDPG,TD3,SAC1}. This mechanism relies on continuous, smooth output changes—a property violated by the non-differentiable, binary nature of SNN spikes. This can cause abrupt output shifts, leading to unstable optimization objective, and oscillatory updates. Such instability not only makes the model highly sensitive to random seed initialization but also hampers convergence and undermines reliability in real-world deployment.

To address this mismatch between discrete spikes and continuous-control updates, we propose a proxy target framework for SNN-based RL (Fig.~\ref{fig:big_fig}(c)). Instead of using an SNN target actor, we introduce a differentiable proxy actor network that imitates the behavior of the online spiking actor network. The proxy target network can alter its output smoothly and continuously, stabilizing the learning process and improving performance, as demonstrated in Fig.~\ref{fig:big_fig}(d). Since the proxy target network is only used for auxiliary training, the proposed approach retains SNN's advantages of low-latency and energy efficiency during inference in real world applications. Our main contributions are summarized as follows:
\begin{itemize}
    \item We identify a critical mismatch between discrete SNN outputs and the continuous target network soft update mechanism used in off-policy RL, showing how this conflict destabilizes training and degrades performance.
    \item We propose a proxy target framework that replaces the spiking target network with a continuous, differentiable proxy, enabling smooth target updates and stable optimization.
    \item We introduce an implicit gradient-based update rule that aligns the proxy with the online SNN, mitigating target output gaps and giving precise optimization goals.
    \item Extensive experiments across multiple neuron models and continuous control benchmarks demonstrate consistent stability improvements and up to $\mathbf{32\%}$ higher average performance. To the best of our knowledge, this is the first approache where SNNs with simple Leaky Integrate-and-Fire (LIF) neurons surpass ANN performance in continuous control.
\end{itemize}

\section{Related works}
\subsection{Learning rules of SNN-based RL}
\paragraph{Synaptic plasticity.} Inspired by the plasticity of biological synapses, several works have integrated SNNs into reinforcement learning via reward-modulated spike-timing-dependent plasticity (R-STDP)~\citep{R-STDP,R-STDP_3factor,R-STDP_eligibity_trace,continuousAC,SVPG}. These approaches are biologically plausible and energy-efficient, but have limited performance on complex tasks.

\paragraph{ANN-SNN conversion.} With the progress of ANN-based deep RL and ANN-SNN conversion algorithms~\citep{cao2015spiking,bu2022optimized,bu2021optimal}, some studies~\citep{DQN2ANN1,DQN2ANN2,DQN_2ANN_BP} convert well-trained Deep Q-Networks (DQNs)~\citep{Atari,Mnih2015HumanlevelCT} into SNNs. Such conversion-based methods achieve lower energy consumption during inference, but require ANN pre-training. 

\paragraph{Gradient-based direct training.} To avoid ANN pre-training, several works~\citep{DQN_BP1,DQN_BP2,BP_DQN_AC,sun2022solving,sun2025multi} directly train SNNs for RL using STBP~\citep{STBP}, while \citet{learningd_deliema} introduced e-prop with eligibility traces to learn policies through the policy gradient algorithm~\citep{policy_gradient}. These approaches achieve competitive results in discrete action spaces, but they cannot be extended to continuous control tasks.

\subsection{Hybrid framework of spiking actor network.}

In continuous-control problems where the action space is continuous, hybrid frameworks have been extensively explored. \citet{SDDPG} first proposed an SNN-based actor co-trained with an ANN critic in the Actor-Critic framework~\citep{Actor_Critic}. \citet{popSAN} demonstrated that population encoding improves the performance of spiking actor networks. Subsequent works enhanced these frameworks through various mechanisms, such as utilizing dynamic neurons \citep{MDC_SAN}, incorporating lateral connections \citep{ILC_SAN}, adding bio-plausible topologies \citep{BPT_SAN}, and integrating dynamic thresholds \citep{dynamic_threshold}. 

While these hybrid approaches report performance comparable to or exceeding their ANN counterparts, two key limitations remain. First, they often rely on complex neuron models (e.g., current-based LIF or second-order dynamic neurons), increasing computational cost and training difficulty. Second, the RL algorithms themselves are not modified to account for SNN-specific dynamics, which may cause instability and suboptimal convergence. In contrast, our proxy target framework is tailored to the discrete, event-driven nature of SNNs, achieving superior stability and performance with only simple LIF neurons.

\section{Preliminaries}
To avoid ambiguity, we use \emph{training steps} to denote RL time steps and \emph{simulation steps} to denote internal SNN simulation time steps.

\subsection{Reinforcement Learning}
Reinforcement Learning (RL) involves an agent interacting with an environment. The agent observes the current state $s$, performs an action $a$ and receives a reward $r$, while the environment transitions to the next state $s'$. The agent’s objective is to learn a policy $\pi_\phi$, parameterized by $\phi$, that maximizes the expected return.

In continuous control settings, the action space is a continuous vector (e.g., torque values). Most continuous control algorithms adopt the Actor–Critic framework with a deterministic policy~\citep{RL1}, where the actor $\pi_\phi$ outputs actions $a=\pi_\phi(s)$ and the critic $Q_\theta$ evaluates them with parameters $\theta$~\citep{Actor_Critic}. The actor is updated by the deterministic policy gradient~\citep{DPG}:
\begin{equation}
    \label{eq:actor_update}
    \nabla_\phi J(\phi)=\mathbb{E}\left[\nabla_aQ_\theta(s,a)\mid_{a=\pi(s)}\nabla_\phi\pi_\phi(s)\right].
\end{equation}
The critic is updated via temporal-difference (TD) learning~\citep{TD} using the Bellman equation~\citep{Bellman}:
\begin{equation}
\label{eq:critic_update}
    Q_\theta(s,a)\gets y, \quad y=r+\gamma Q_{\theta'}(s',a'),\space\space a'=\pi_{\phi'}(s'),    
\end{equation}
where $\gamma$ is the discount factor and $(\pi_{\phi'}, Q_{\theta'})$ denote target networks.

\subsection{Target network soft update}
The target networks $(\pi_{\phi'}, Q_{\theta'})$ share the same architecture as their online counterparts $(\pi_\phi, Q_\theta)$ but are updated more slowly to provide stable learning targets. Their parameters are updated by the Polyak function with smoothing factor $\tau$:
\begin{equation}
    \label{eq:target_update}
    \begin{array}{c}
        \phi'\gets\tau \phi+(1-\tau)\phi',\quad \theta'\gets\tau \theta+(1-\tau)\theta'.
    \end{array}
\end{equation}
These soft updates play a crucial role in off-policy continuous-control algorithms. As shown in Eqs.~(\ref{eq:actor_update}, \ref{eq:critic_update}), the actor and critic are jointly optimized through bootstrapping, which can cause oscillatory updates due to their mutual dependence. The target networks mitigate this by producing slowly changing targets, thereby stabilizing training and preventing divergence.

\subsection{Spiking Neural Networks}
\label{sec:SNN}
\paragraph{Spiking neuron model.} In an SNN, each neuron integrates presynaptic spikes into its membrane potential and emits a spike when the potential exceeds a threshold. The Leaky Integrate and Fire (LIF) neuron~\citep{LIF} is one of the most widely used models, governed by the following dynamics:
\begin{align}
    I_t^l &= W^l S_t^{l-1} + b^l, \label{eq:current} \quad H_t^l = \lambda V_{t-1}^l + I_t^l,\\
    S_t^l &= \Theta(H_t^l - V_{th}),  \quad  V_t^l = (1 - S_t^l) H_t^l + S_t^l \cdot V_{\text{reset}},
\end{align}
where $I$ is the input current, $H$ is the accumulated membrane potential, $S$ is the binary output spike, $V$ is the membrane potential after the firing process. $W$ and $b$ are the weights and the biases, $V_{th}$, $V_{\text{reset}}$, and $\lambda$ are the threshold voltage, the reset voltage and the membrane leakage parameter, respectively. All subscripts $(\cdot)_t$ and all superscripts $(\cdot)^l$ denote simulation step $t$ and layer $l$ respectively. $\Theta(\cdot)$ is the Heaviside function.

\paragraph{Spiking actor network.} The spiking actor network (SAN) consists of a population encoder with Gaussian receptive fields \citep{popSAN}, a multi-layer SNN, and a decoder that uses the membrane potentials of non-firing neurons as continuous outputs \citep{ILC_SAN}. The SAN is trained using STBP with a surrogate gradient function. Detailed forward and backward formulations are provided in Appendix~\ref{APP:SAN}.

\section{Methodology}

In this section, we propose a novel proxy target framework to address the incompatibility between the discrete dynamics of spiking neurons and the continuous target network soft update mechanism in RL. Section~\ref{sec:discrete_target} analyzes the instability caused by discrete target outputs and introduces a proxy target network with continuous dynamics. Section~\ref{sec:address_error} presents an implicit imitation mechanism that aligns the proxy network with the online SNN through gradient-based optimization. Section~\ref{sec:overall_algo} summarizes the overall training procedure.

\begin{figure}[htbp]
  \centering
\includegraphics[width=1\linewidth]{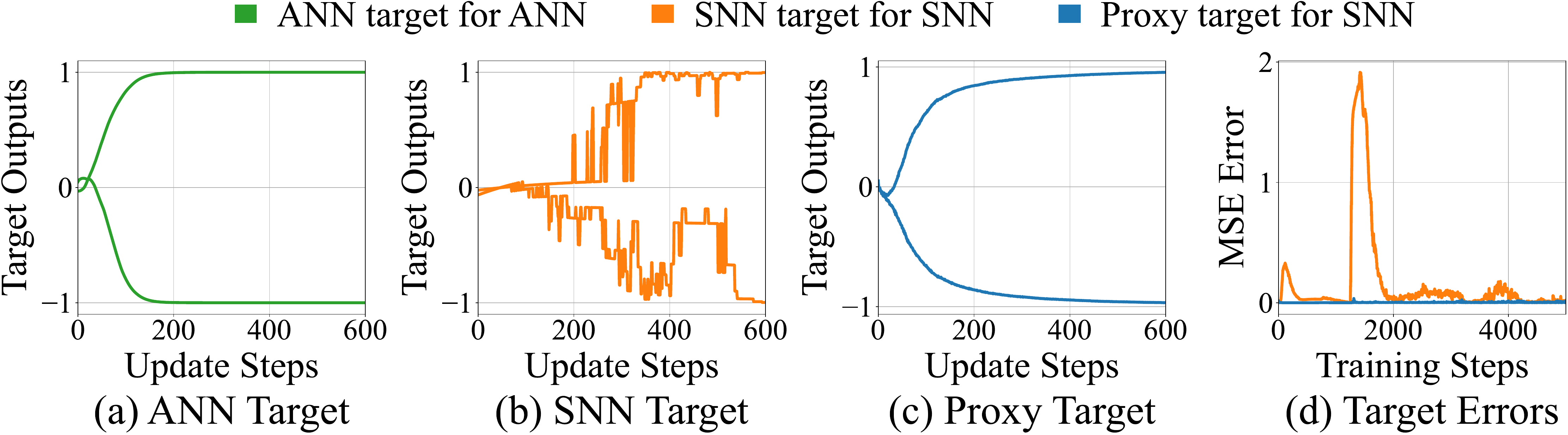}
  \caption{Effects of different target network update mechanisms. (a)-(c) show output trajectories of different target networks during updates, where each line denotes a normalized output dimension within $(-1,1)$. (a) ANN target network exhibits smooth transitions; (b) SNN target network produces discrete and irregular output jumps; (c) the proposed proxy target achieves continuous and stable transitions. (d) Mean squared error between target and online networks during training in the InvertedDoublePendulum-v4 environment.}
  \label{fig:proxy_effect}
\end{figure}

\subsection{Addressing discrete targets by proxy network}
\label{sec:discrete_target}
\paragraph{Performance degradation due to discrete target outputs.}
In the standard Actor–Critic framework, the target network is updated using the Polyak function (Eq.~\ref{eq:target_update}), which assumes that small parameter updates lead to smooth output transitions. This assumption holds for ANNs with continuous activation functions but fails for SNNs, whose firing function is binary and non-differentiable. To illustrate this effect, we construct target networks corresponding to trained online networks using identical architectures and neuron models. The target parameters are updated according to Eq.~\ref{eq:target_update} with $\tau = 0.005$ (the most commonly used setting), while the online network is frozen. Figures~\ref{fig:proxy_effect}(a)–(b) show the target outputs during updates: the ANN target evolves smoothly, whereas the SNN target exhibits frequent discontinuous jumps. Although both targets eventually converge to their online counterparts, the discrete shifts in the SNN target (Fig.~\ref{fig:proxy_effect}(b)) cause erratic transitions that propagate instability to the critic’s optimization objectives, resulting in oscillatory and unreliable learning dynamics~\citep{TD3}.

\paragraph{Smoothing target outputs by proxy network.}
As illustrated in Fig.~\ref{fig:big_fig}(c), to restore smoothness, we introduce a proxy target network that replaces the discrete spiking neurons of SNNs with continuous activation functions of ANNs. As shown in Fig.~\ref{fig:proxy_effect}(c), the proxy network produces gradual output transitions during updates, effectively eliminating the discrete jumps observed in SNN targets.  
This design enables stable soft updates and prevents abrupt shifts in the target outputs, thereby improving the stability of the overall Actor–Critic learning process.

\subsection{Addressing target output gaps by implicit updates}
\label{sec:address_error}
\paragraph{Performance degradation due to target output gaps.} 

Although the proxy network achieves smooth updates, directly substituting spiking neurons with continuous activations (e.g., ReLU) introduces an output gap between the proxy and the online SNN. This discrepancy prevents the proxy target from accurately reproducing the output of the online SNN, distorting the critic’s learning targets and reducing overall policy performance.

\begin{figure}[htbp]
  \centering
  \includegraphics[width=1\linewidth]{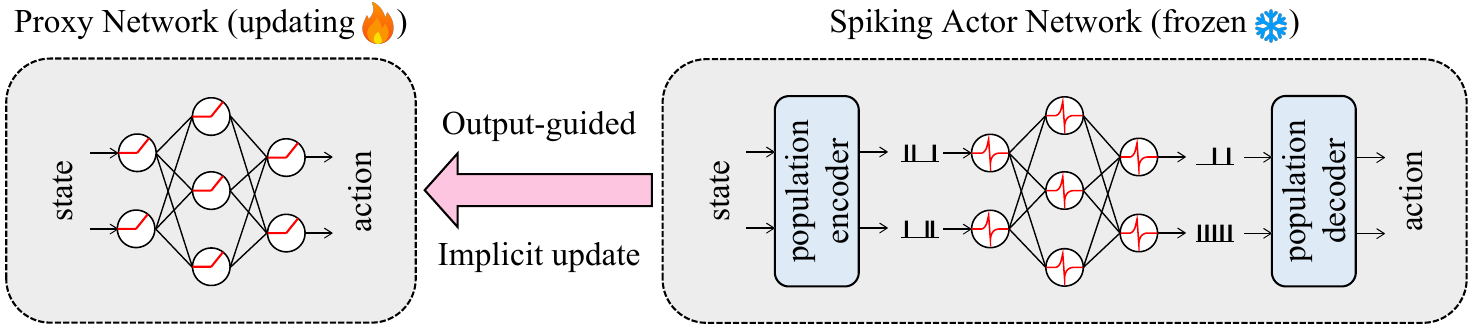}
  \caption{Architecture of the proposed proxy network and the spiking actor network. The proxy actor is updated implicitly by imitating the behavior of the online spiking actor network, ensuring stable and accurate target updates.}
  \label{fig:network_arch}
\end{figure}

\paragraph{Aligning proxy network by implicit updates.} 

Since the approximation errors cannot be eliminated by explicitly copying the weights of the online SNN, we propose an implicit proxy update method. As shown in Fig.~\ref{fig:network_arch}, unlike the explicit soft update that directly averages parameters, our approach computes updates in the output space, which gradually reduces the gap between the online SNN and the proxy target. Let the proxy actor $\pi_{\phi'}^{\text{Proxy}}$ have parameters $\phi'$ and the online spiking actor $\pi_{\phi}^{\text{SNN}}$ have parameters $\phi$. For each input state $s$, the proxy output is implicitly updated toward the SNN actor output as:
\begin{equation}
    \label{eq:output_match}
    \pi_{\phi'}^\text{Proxy}(s)\gets (1-\tau')\cdot\pi_{\phi'}^\text{Proxy}(s)+\tau' \cdot \pi_{\phi}^\text{SNN}(s).
\end{equation}
where $\tau'$ is a smoothing coefficient similar to $\tau$ in Eq.~\ref{eq:target_update}. Since it is difficult to directly update the corresponding parameter according to Eq.~\ref{eq:output_match}, we instead perform a gradient-based optimization that achieves a similar effect:
\begin{equation}
\phi' \gets \phi' + \tau \left(\pi_{\phi}^\text{SNN}(s)-\pi_{\phi'}^\text{Proxy}(s) \right)\nabla_{\phi'}\pi_{\phi'}^\text{Proxy}(s) =\phi'-\frac{\tau}{2} \nabla_{\phi'}  \left\| \pi_{\phi'}^\text{Proxy}(s)) - \pi_{\phi}^\text{SNN}(s))\right\|_2^2,
 \label{eq:}
\end{equation}
where $\|\cdot\|_2^2$ denotes the squared $\ell_2$ norm. Thus, the proxy network can be updated by gradient descent that minimizes the proxy loss:
\begin{equation}
    \label{eq:BC}
    L_{proxy}=\frac{1}{N} \sum_{i=1}^{N} \left\| \pi_{\phi'}^\text{Proxy}(s_i)) - \pi_{\phi}^\text{SNN}(s_i))\right\|_2^2,
\end{equation}
where $N$ denotes the batch size, $s_i$ are the states sampled from the replay buffer in RL algorithm. This proxy update mechanism acts as a form of implicit imitation learning, aligning the proxy network with the SNN actor while maintaining smooth output transitions, as demonstrated in Theorem~\ref{Thm:1}.

\newtheorem{Theorem}{Theorem}
\begin{Theorem}
\label{Thm:1}
Let the proxy network $\pi_{\phi'}^{\text{Proxy}}$ be updated by minimizing the loss $L_{\text{proxy}}$ in Eq.~\ref{eq:BC}.  
During each update, as the proxy learning rate $lr_{\text{proxy}} \to 0$, the output change satisfies
\[
\|\pi_{\phi'_{\text{new}}}^{\text{Proxy}}(s)
- \pi_{\phi'_{\text{old}}}^{\text{Proxy}}(s)\| \to 0,
\]
where $\phi'_{\text{old}}$ and $\phi'_{\text{new}}$ denote parameters before and after the update, respectively. Hence, minimizing $L_{\text{proxy}}$ ensures sufficiently small and smooth policy updates, promoting stable optimization.
\end{Theorem}

Since the proxy network is a multi-layer feedforward model, a universal approximator~\citep{universal_approximate}, it can asymptotically match the SNN actor’s output by minimizing Eq.~\ref{eq:BC}. To further demonstrate this empirically, Fig.~\ref{fig:proxy_effect}(d) shows the mean-squared output gap between the proxy and the SNN actor during training. While the SNN target occasionally diverges from the online SNN, the proxy network remains well-aligned throughout, validating that the proposed approach effectively mitigates target output gaps and provides precise and stable optimization goals for RL training.

\subsection{Overall training framework}
\label{sec:overall_algo}

\begin{algorithm}[htbp]
    \caption{Proxy Target framework}
    \label{algo:general}
    \begin{algorithmic}[1]
        \STATE Initialize SNN actor network $\pi_\phi^{\text{SNN}}(s)$, ANN critic network $Q_\theta^{\text{ANN}}(s,a)$ with parameters $\phi$, $\theta$
        \STATE Initialize proxy actor $\pi_{\phi'}^{\text{Proxy}}(s)$ and ANN target critic $Q_{\theta'}^{\text{ANN}}(s,a)$ with parameters $\phi'$ and $\theta'$
        \STATE Initialize replay buffer $\mathcal{D}$
        \FOR{each iteration}
        \STATE Execute action $a$ according to $\pi_\phi^{\text{SNN}}(s)$ and store the transition $(s, a, r, s')$ in $\mathcal{D}$
        \IF{proxy target update}
        \FOR{$k=1$ to $K$}
            \STATE Sample a minibatch of $N$ transitions $(s_i, a_i, r_i, s'_i)$ from $\mathcal{D}$
            \STATE Update proxy actor parameters $\phi'$ by minimizing:$$L_{proxy}=\frac{1}{N} \sum_i \left\| \pi_{\phi'}^{\text{Proxy}}(s_i) - \pi_\phi^{\text{SNN}}(s_i) \right\|_2^2$$
        \ENDFOR
        \ENDIF
        \IF{ANN target update}
        \STATE Update ANN target critic parameters $\theta'$ by the Polyak function: $\theta'\gets\tau\theta + (1-\tau)\theta'$
        \ENDIF
            
        \IF{ANN critic update}
        \STATE Compute target values $y_i$ using proxy actor $\pi_{\phi'}^{\text{Proxy}}$ and target critic $Q_{\theta'}^{\text{ANN}}$
        \STATE Update ANN critic by minimizing: $L_{critic}=\frac{1}{N}\sum_i\left(Q_\theta^{\text{ANN}}(s_i,a_i)-y_i\right)^2$
        \ENDIF

        \IF{SNN actor update}
            \STATE Update SNN actor by maximizing: $J=\frac{1}{N}\sum_iQ_\theta^{\text{ANN}}\left(s_i,\pi_\phi^{\text{SNN}}(s_i)\right)$
        \ENDIF
        
        \ENDFOR
    \end{algorithmic}
\end{algorithm}

The proposed proxy target framework is shown in Fig.~\ref{fig:big_fig}(c). The proxy actor network contains continuous activations of ANN that replace the discontinuous SNN target actor network. Instead of explicitly updating network parameters, the proxy actor is implicitly optimized to imitate the behavior of the online SNN actor by minimizing the loss in Eq.~\ref{eq:BC}. During each update episode, the proxy actor is optimized for $K$ iterations to reliably approximate the discrete SNN outputs, compensating for the greater representational difficulty. Meanwhile, the target critic is updated explicitly by copying weights using the Polyak function, as both the critic and target critic are conventional ANNs. The complete training procedure is summarized in Algorithm~\ref{algo:general}.

As demonstrated in Theorem~\ref{Thm:1} and Fig.~\ref{fig:proxy_effect}(c)-(d), the proxy actor not only produces smooth output transitions but also closely tracks the SNN actor’s behavior\footnote[1]{Minor fluctuations in the proxy network output (Fig.~\ref{fig:proxy_effect}(c)) resemble the stochasticity introduced by soft target updates with noise injection in DRL~\citep{TD3}, which can further reduce overfitting in value estimation.}. The proxy target framework effectively alleviates the instability caused by discrete and imprecise targets in traditional SNN–RL frameworks, resulting in a more stable training process within the Actor–Critic framework.

It is worth noting that the proposed mechanism preserves the energy efficiency of SNNs, as the proxy network and the critic network are used exclusively during training and are discarded during deployment, introducing no additional computational overhead during deployment.   

\section{Experiments}
\subsection{Experimental setup}
\begin{figure}[htbp]
  \centering
\includegraphics[width=0.8\linewidth]{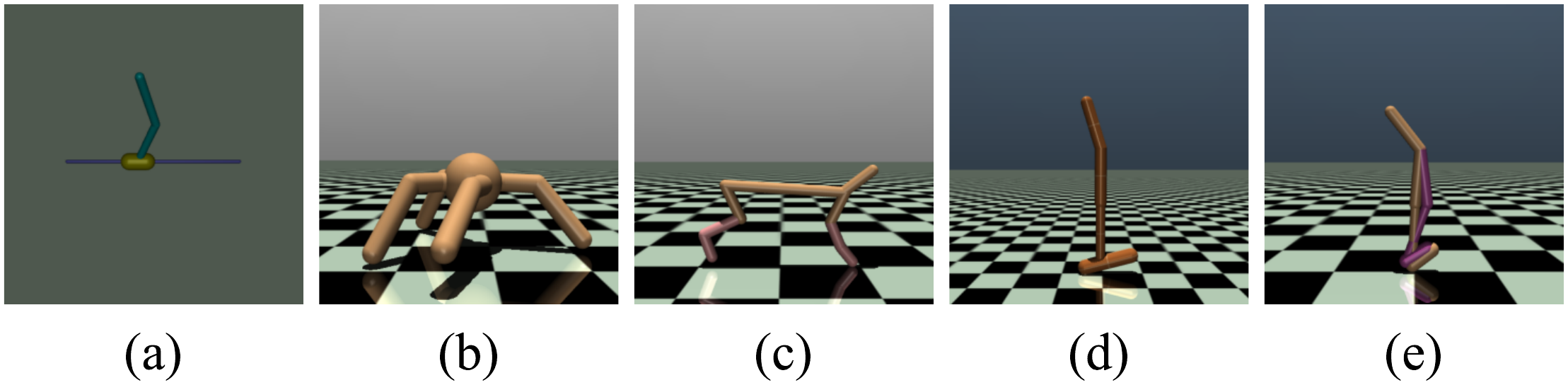}
  \caption{Continuous control tasks of the MuJoCo environments on OpenAI Gymnasium. (a) InvertedDoublePendulum-v4, (b) Ant-v4, (c) HalfCheetah-v4, (d) Hopper-v4, (e) Walker2d-v4.}
  \label{fig:envs}
\end{figure}
The proposed proxy target framework (PT) was evaluated across multiple continuous-control tasks in the MuJoCo simulator~\citep{mujoco1,mujoco2} using the OpenAI Gymnasium benchmark suite~\citep{gym,gymnasium}, including \texttt{InvertedDoublePendulum-v4} (IDP)~\citep{IDP}, \texttt{Ant-v4}~\citep{GAE}, \texttt{HalfCheetah-v4}~\citep{halfcheetah}, \texttt{Hopper-v4}~\citep{hopper}, and \texttt{Walker2d-v4}. All environments follow the default configurations without modifications.  

The experiments were carried out with different spiking neuron models, such as the LIF neuron, the current-based LIF neuron (CLIF)~\cite{popSAN}, and the dynamic neuron (DN)~\cite{MDC_SAN}. The LIF and CLIF neuron parameters follow~\cite{popSAN}, while the DN parameters are initialized as in~\cite{MDC_SAN}.

We tested the proposed algorithm in conjunction with the TD3 algorithm \citep{TD3}, all detailed parameter settings are provided in Appendix \ref{app:exp_para}. For a fair comparison, all spiking actor networks share the same architecture, encoding, and decoding schemes provided in Appendix \ref{APP:SAN}. All SNNs have a simulation step of $\mathbf{5}$ unless otherwise noted. All reported data in this section are reproduced results across five random seeds.

\subsection{Results across different spiking neurons} 
\begin{figure}[htbp]
  \centering
\includegraphics[width=1\linewidth]{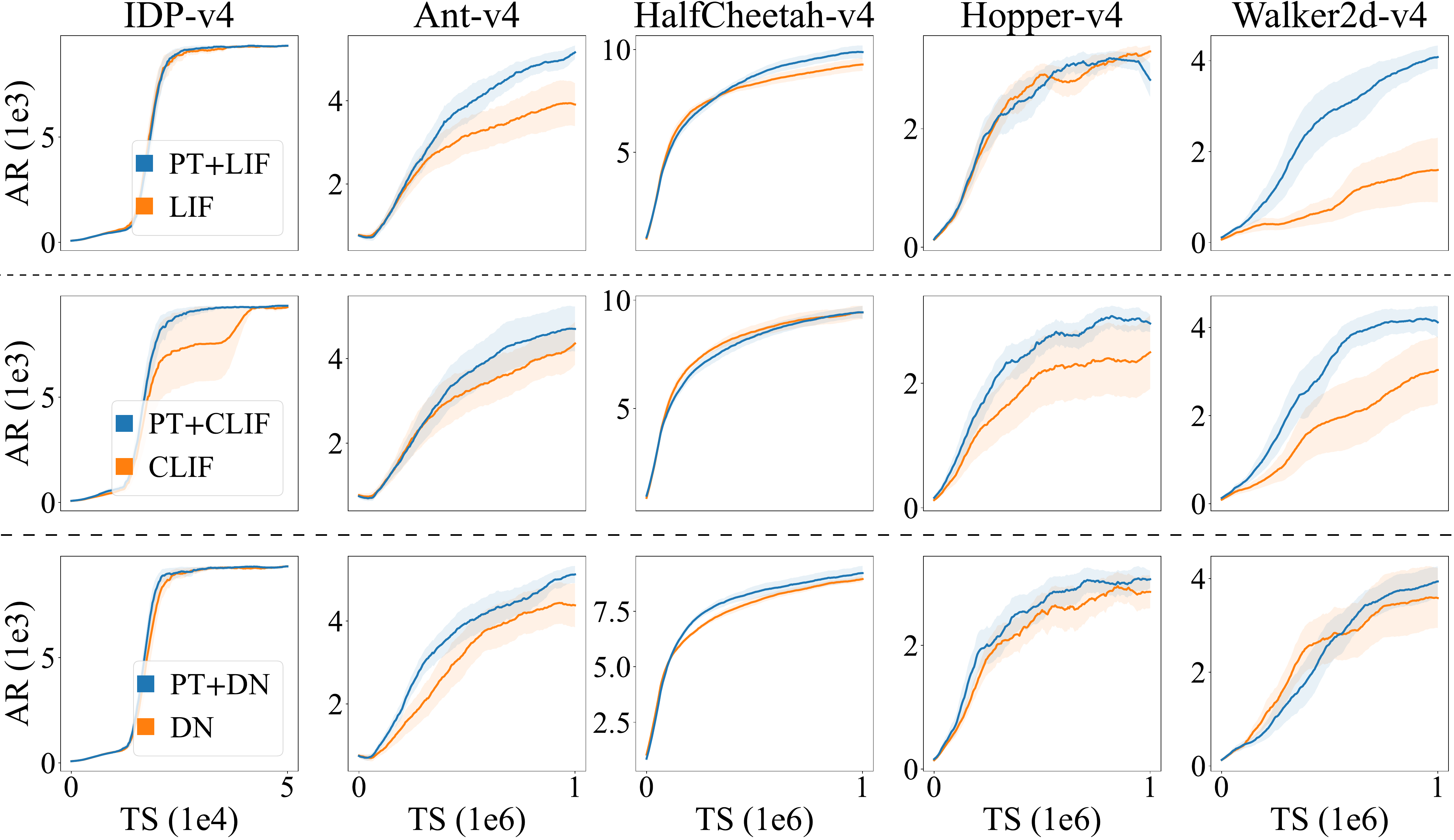}
  \caption{Learning curves of the proxy target framework (PT) and the vanilla Actor–Critic framework with the LIF neuron, the CLIF neuron and the DN. AR denotes average returns, and TS denotes training steps. The shaded region represents half a standard deviation over 5 different seeds. Curves are uniformly smoothed for visual clarity.}
  \label{fig:learning_curves}
\end{figure}
\paragraph{Increasing performance.} Fig.~\ref{fig:learning_curves} shows the learning curves of the proposed proxy target framework and the vanilla Actor-Critic framework with different spiking neurons. The proxy target framework improves the performance of different spiking neurons, demonstrating its general applicability in delivering both faster convergence and higher final returns across different neuron types and environments.  
\begin{figure}[htbp]
  \centering
\includegraphics[width=1\linewidth]{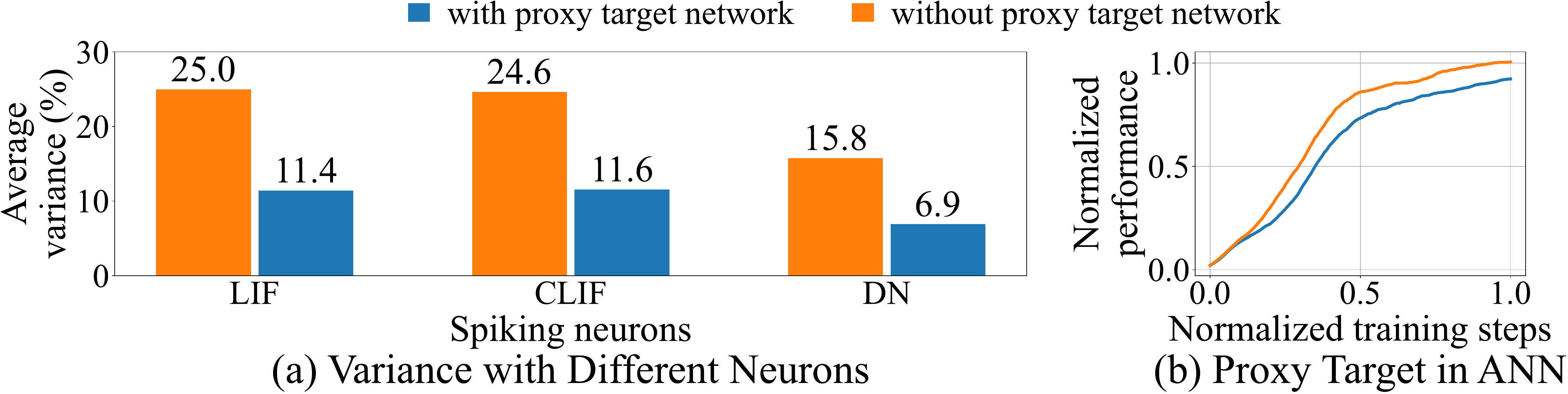}
  \caption{(a) Average variance of different neurons after training. The average variance is computed by averaging the standard deviation ratio with $5$ seeds, across all environments. (b) Normalized learning curves across all environments of the ANN integrated with the proposed proxy network across all environments. The performance and training steps are normalized linearly to $(0,1)$. Curves are uniformly smoothed for visual clarity.}
  \label{fig:var_ablation}
\end{figure}
\paragraph{Improving stability.} Fig.~\ref{fig:var_ablation}(a) shows the performance variance (after training) of the proxy target framework and the vanilla Actor-Critic framework with different spiking neurons. The proxy target framework reduces the variance of different spiking neurons, demonstrating its capability to stabilize training. This is crucial for real‐world deployments, where retraining costs are high and consistent behavior is required.

\subsection{Exceeding state-of-the-art} 
To quantify relative improvements, we define the average performance gain (APG) as:
\begin{equation}
    \label{eq:APR}
    APG=\left(\frac{1}{|\text{envs}|}\sum_{\text{env}\in \text{envs}}\frac{\text{performance}(\text{env})}{\text{baseline}(\text{env})}-1\right)\cdot100\%,
\end{equation}

where $|\text{envs}|$ denotes the total number of environments, $\text{performance}(\text{env})$ and $\text{baseline}(\text{env})$ are the performance of the algorithm and the baseline in that particular environment. Tab.\ref{tab:cmp_sota} compares our proxy target framework with ANN-based RL, the ANN–SNN conversion method~\citep{bu2025inference} (100 simulation steps), and other state-of-the-art SNN-based RL algorithms, including pop-SAN \citep{popSAN}, MDC-SAN \citep{MDC_SAN}, and ILC-SAN \citep{ILC_SAN}. With the proxy network, a simple LIF-based SNN surpasses all baselines, including those using complex neuron dynamics or connection structures, and achieves higher average returns than standard ANNs. Although performance varies across tasks, the average gain across all environments and neurons indicates the general applicability of the proxy target framework.  

\begin{table}[htbp]
  \footnotesize
  \caption{Max average returns over $5$ random seeds with different spiking neurons, and the average performance gain against the ANN baseline, where $\pm$ denotes one standard deviation.}
  \label{tab:cmp_sota}
  \centering
  \resizebox{1.0\textwidth}{!}{
  \begin{tabular}{lcccccc}
    \toprule
    Method&IDP-v4&Ant-v4&HalfCheetah-v4&Hopper-v4&Walker2d-v4& APG\\ \midrule
      ANN (TD3)& $7503 \pm 3713$& $4770 \pm 1014$&$10857 \pm 475$& $3410 \pm 164$& $4340 \pm 383$& $0.00\%$\\
 ANN-SNN&$3859\pm4440$&$3550\pm963$&$8703\pm658$&$3098\pm281$&$4235\pm354$&$-21.11\%$\\
 Vanilla LIF& $9347 \pm 1$& $4294 \pm 1170$& $9404 \pm 625$& $3520 \pm 94$& $1862 \pm 1450$&$-10.54\%$\\
 pop-SAN& $9351 \pm 1$& $4590 \pm 1006$& $9594 \pm 689$& $2772 \pm 1263$& $3307 \pm 1514$&$-6.66\%$\\
 MDC-SAN& $9350 \pm 1$& $4800 \pm 994$& $9147 \pm 231$& $3446 \pm 131$& $3964 \pm 1353$&$0.37\%$\\
 ILC-SAN& $9352 \pm 1$& $5584 \pm 272$& $9222 \pm 615$& $3403 \pm 148$& $4200 \pm 717$&$4.64\%$\\
 \midrule
 PT-CLIF& $9351 \pm 1$& $5014 \pm 1074$& $9663 \pm 426$& $3526 \pm 112$& $4564 \pm 555$&$5.46\%$\\
 PT-DN& $9350 \pm 1$& $5400 \pm 277$& $9347 \pm 666$& $3507 \pm 144$& $4277 \pm 650$&$5.06\%$\\
         PT-LIF&$9348 \pm 1$&$5383 \pm 250$&$10103 \pm 607$&$3385 \pm 157$&$4314 \pm 423$&$\mathbf{5.84\%}$\\
    % \midrule
    \bottomrule
  \end{tabular}
  }
\end{table}

\subsection{Simple neurons perform best} 
Interestingly, the simplest LIF neuron achieves the highest overall performance under the proxy target framework.  
This contrasts with previous findings where complex neuron models generally perform better.  
Once the SNN surpasses its ANN counterpart, the primary performance bottleneck shifts from the neuron model to the RL algorithm itself.  
Hence, introducing more complex spiking dynamics may unnecessarily increase training difficulty and even degrade performance.

\subsection{SNN-friendly design} 
Fig.~\ref{fig:var_ablation}(b) shows the normalized performance of ANN with and without the proxy network. The proxy target framework cannot improve the performance in ANNs, confirming that the observed benefits arise from addressing SNN-specific challenges rather than providing a stronger RL algorithm. This validates the SNN-friendly design of our framework.

\subsection{Energy efficiency} 
Finally, we evaluate inference energy consumption across models. The comparison includes a traditional ANN-based TD3 model, a baseline spiking actor network using vanilla LIF neurons, and our PT-LIF model. The consumption is estimated as the same way as \cite{merolla2014million}, where multiply-accumulate (MAC) operation costs $3.97$pJ on modern NPUs\footnote{Energy per MAC is estimated from the maximum inference throughput (mJ$^{-1}$) across seven NPUs: MAX78k (C/R), GAP8, NXP-MCXN947, HX-WE2 (S/P), and MILK-V~\citep{NPU}. NPU initialization energy is excluded.}~\citep{NPU} and synaptic operation (SOP) costs $77$fJ~\citep{hu2021spiking}. 
\begin{table}[htbp]
  \footnotesize
  \caption{Energy consumptions of different tasks per inference for the spiking actor network with LIF neurons, where the energy unit is nano-joule (nJ).}
  \label{tab:energy}
  \centering
  \resizebox{1.0\textwidth}{!}{
  \begin{tabular}{lcccccc}
    \toprule
    Method&IDP-v4&Ant-v4&HalfCheetah-v4&Hopper-v4&Walker2d-v4& Average\\ \midrule
      ANN (TD3)& $72.33$&$295.60$&$283.41$&$274.26$&$283.41$&$281.78$ ($71014.4$ MACs)\\
 Vanilla LIF& $8.14$& $11.78$& $15.13$& $7.21$& $18.82$&$12.21$ ($158.6\times 10^3$ SOPs)\\
         PT-LIF&$9.01$&$12.18$&$13.46$&$6.86$&$13.93$&$\mathbf{11.09}$ (${144.0\times 10^3}$ SOPs)\\
    \bottomrule
  \end{tabular}
  }
\end{table}

As shown in Tab.\ref{tab:energy}, the ANN (TD3) model consumes significantly more energy, while both spiking models demonstrate dramatically lower energy consumption. Specifically, our proposed PT-LIF model achieves the lowest average consumption while maintaining better stability and performance. Moreover, PT-LIF's average firing rate (32\%) is slightly lower than that of the vanilla LIF model (33\%), further improving energy efficiency. These results highlight the superior energy efficiency of the proposed method, making it compelling for deployment on energy-constrained platforms.

\section{Conclusion}
In this work, we identified a critical mismatch between the discrete dynamics of SNNs and the continuous requirement of the target network soft update mechanism in the Actor-Critic framework. To address this, we proposed a novel proxy target framework that enables smooth target updates and faster convergence. Experimental results demonstrate that the proxy network can stabilize training and improve performance, enabling simple LIF neurons to surpass ANN performance in continuous control. 

In contrast to previous works which retrofit SNNs into ANN-centric RL frameworks, this work opens a door to investigate and design SNN-friendly RL algorithms which is tailored to SNN's specific dynamics. In the future, more SNN-specific adjustments could be applied to SNN-based RL algorithms to improve performance and energy efficiency in real-world, resource-constrained RL applications. 

\textbf{Limitation.} While this work designs a proxy target framework that is suitable for SNN-based RL, it still remains at the simulation level. The next step may involve implementing it on edge devices and enabling decisions-making in the real world. 

\section{Acknowledgement}
This work is funded by National Natural Science Foundation of China (62422601, U24B20140, and  62088102), Beijing Municipal Science and Technology Program (Z241100004224004) and Beijing Nova Program (20230484362, 20240484703), and National Key Laboratory for Multimedia Information Processing.

\bibliographystyle{plainnat}
\bibliography{arbv.bib,refs.bib}

\section*{NeurIPS Paper Checklist}

\begin{enumerate}

\item {\bf Claims}
    \item[] Question: Do the main claims made in the abstract and introduction accurately reflect the paper's contributions and scope?
    \item[] Answer: \answerYes{} % Replace by \answerYes{}, \answerNo{}, or \answerNA{}.
    \item[] Justification: In the abstract and introduction sections, we clearly point out the contributions of this work.

\item {\bf Limitations}
    \item[] Question: Does the paper discuss the limitations of the work performed by the authors?
    \item[] Answer: \answerYes{}{} % Replace by \answerYes{}, \answerNo{}, or \answerNA{}.
    \item[] Justification: We point out the limitations in the last section.

\item {\bf Theory assumptions and proofs}
    \item[] Question: For each theoretical result, does the paper provide the full set of assumptions and a complete (and correct) proof?
    \item[] Answer: \answerNA{} % Replace by \answerYes{}, \answerNo{}, or \answerNA{}.
    \item[] Justification: This work does not include theoretical results.

    \item {\bf Experimental result reproducibility}
    \item[] Question: Does the paper fully disclose all the information needed to reproduce the main experimental results of the paper to the extent that it affects the main claims and/or conclusions of the paper (regardless of whether the code and data are provided or not)?
    \item[] Answer: \answerYes{} % Replace by \answerYes{}, \answerNo{}, or \answerNA{}.
    \item[] Justification: We provide detailed experimental setups in the Appendix.

\item {\bf Open access to data and code}
    \item[] Question: Does the paper provide open access to the data and code, with sufficient instructions to faithfully reproduce the main experimental results, as described in supplemental material?
    \item[] Answer: \answerYes{} % Replace by \answerYes{}, \answerNo{}, or \answerNA{}.
    \item[] Justification: We provide codes with sufficient instructions in the supplementary materials. 

\item {\bf Experimental setting/details}
    \item[] Question: Does the paper specify all the training and test details (e.g., data splits, hyperparameters, how they were chosen, type of optimizer, etc.) necessary to understand the results?
    \item[] Answer: \answerYes{} % Replace by \answerYes{}, \answerNo{}, or \answerNA{}.
    \item[] Justification: We show all setups and hyper-parameters in the Appendix.

\item {\bf Experiment statistical significance}
    \item[] Question: Does the paper report error bars suitably and correctly defined or other appropriate information about the statistical significance of the experiments?
    \item[] Answer: \answerYes{} % Replace by \answerYes{}, \answerNo{}, or \answerNA{}.
    \item[] Justification: The learning curves figure and the main result table show the standard deviations. 

\item {\bf Experiments compute resources}
    \item[] Question: For each experiment, does the paper provide sufficient information on the computer resources (type of compute workers, memory, time of execution) needed to reproduce the experiments?
    \item[] Answer: \answerYes{} % Replace by \answerYes{}, \answerNo{}, or \answerNA{}.
    \item[] Justification: We give the experiments compute resources in the Appendix.
    
\item {\bf Code of ethics}
    \item[] Question: Does the research conducted in the paper conform, in every respect, with the NeurIPS Code of Ethics \url{https://neurips.cc/public/EthicsGuidelines}?
    \item[] Answer: \answerYes{} % Replace by \answerYes{}, \answerNo{}, or \answerNA{}.
    \item[] Justification: Our work conform with the NeurIPS Code of Ethics.

\item {\bf Broader impacts}
    \item[] Question: Does the paper discuss both potential positive societal impacts and negative societal impacts of the work performed?
    \item[] Answer: \answerNA{} % Replace by \answerYes{}, \answerNo{}, or \answerNA{}.
    \item[] Justification: We think there is no societal impact to be emphasized.
    
\item {\bf Safeguards}
    \item[] Question: Does the paper describe safeguards that have been put in place for responsible release of data or models that have a high risk for misuse (e.g., pretrained language models, image generators, or scraped datasets)?
    \item[] Answer: \answerNA{} % Replace by \answerYes{}, \answerNo{}, or \answerNA{}.
    \item[] Justification: We think there is no such risks.

\item {\bf Licenses for existing assets}
    \item[] Question: Are the creators or original owners of assets (e.g., code, data, models), used in the paper, properly credited and are the license and terms of use explicitly mentioned and properly respected?
    \item[] Answer: \answerYes{} % Replace by \answerYes{}, \answerNo{}, or \answerNA{}.
    \item[] Justification: We cite the original papers properly. 

\item {\bf New assets}
    \item[] Question: Are new assets introduced in the paper well documented and is the documentation provided alongside the assets?
    \item[] Answer: \answerNA{} % Replace by \answerYes{}, \answerNo{}, or \answerNA{}.
    \item[] Justification: This work does not release new assets.

\item {\bf Crowdsourcing and research with human subjects}
    \item[] Question: For crowdsourcing experiments and research with human subjects, does the paper include the full text of instructions given to participants and screenshots, if applicable, as well as details about compensation (if any)? 
    \item[] Answer: \answerNA{} % Replace by \answerYes{}, \answerNo{}, or \answerNA{}.
    \item[] Justification: This work dose not involve crowdsourcing nor research with human subjects.

\item {\bf Institutional review board (IRB) approvals or equivalent for research with human subjects}
    \item[] Question: Does the paper describe potential risks incurred by study participants, whether such risks were disclosed to the subjects, and whether Institutional Review Board (IRB) approvals (or an equivalent approval/review based on the requirements of your country or institution) were obtained?
    \item[] Answer: \answerNA{} % Replace by \answerYes{}, \answerNo{}, or \answerNA{}.
    \item[] Justification: This work does not involve crowdsourcing nor research with human subjects.

\item {\bf Declaration of LLM usage}
    \item[] Question: Does the paper describe the usage of LLMs if it is an important, original, or non-standard component of the core methods in this research? Note that if the LLM is used only for writing, editing, or formatting purposes and does not impact the core methodology, scientific rigorousness, or originality of the research, declaration is not required.
    %this research? 
    \item[] Answer: \answerNA{} % Replace by \answerYes{}, \answerNo{}, or \answerNA{}.
    \item[] Justification: All important and original component dose not involve LLM.

\end{enumerate}

\newpage
\appendix

\section{Appendix}
\subsection{Proof of Theorem 1}
\newtheorem{Theorem2}{Theorem}
\begin{Theorem2}
Let the proxy network $\pi_{\phi'}^{\text{Proxy}}$ be updated by minimizing the loss $L_{\text{proxy}}$ in Eq.~\ref{eq:BC}.  
During each update, as the proxy learning rate $lr_{\text{proxy}} \to 0$, the output change satisfies
\[
\|\pi_{\phi'_{\text{new}}}^{\text{Proxy}}(s)
- \pi_{\phi'_{\text{old}}}^{\text{Proxy}}(s)\| \to 0,
\]
where $\phi'_{\text{old}}$ and $\phi'_{\text{new}}$ denote parameters before and after the update, respectively. Hence, minimizing $L_{\text{proxy}}$ ensures sufficiently small and smooth policy updates, promoting stable optimization.
\end{Theorem2}

\newtheorem{Proof}{Proof}
\begin{Proof}
By standard gradient descent, we have:
$$\lim_{lr_{proxy}\to0}\left\|(\phi_{new}'-\phi_{old}')\right\|=\lim_{lr_{proxy} \to 0} \left \|lr_{proxy} \cdot \nabla_{\phi_{old}'} L_{proxy}\right\|=0.$$
Under the assumption that $\pi_{\phi'}^{Proxy}(s)$ is continuously differentiable with respect to $\phi'$, we apply a first-order Taylor expansion:
$$\lim_{lr_{proxy}\to0}\left\|\pi_{\phi_{new}'}^{Proxy}(s)-\pi_{\phi_{old}'}^{Proxy}(s)\right\|= \lim_{lr_{proxy}\to0}\left\|(\phi_{new}'-\phi_{old}')\nabla_{\phi'}\pi_{\phi_{old}'}^{Proxy}(s)\right\|=0.$$
\end{Proof}

\subsection{Spiking actor network architecture}
\label{APP:SAN}

The spiking actor network (SAN) consists of a population encoder with Gaussian receptive fields, a multi-layer SNN with population output, and a decoder with non-firing neurons.
\subsubsection{Forward propagation of the SAN}
In the state encoder, each input dimension consists of $N_{\text{in}}$ soft reset IF neurons with different Gaussian receptive fields with trainable parameters $\mu$ and $\sigma$. The neurons receive a stimulation $A_E$ at every time step and outputs spikes $S^{in}$ according to:
\begin{equation}
    \label{eq:encoder_gaussian}
    A_E = \exp\left[-\frac{1}{2}\frac{(s-\mu)^2}{\sigma^2}\right],
\end{equation}
\begin{equation}
\begin{array}{c}
V_t^{in} = V_{t-1}^{in}- S_{t-1}^{in} + A_E,\\
S_t^{in} = \Theta(V_t^{in} - V_E),
\end{array}
\end{equation}
where $V_E$ is the threshold of the encoding populations.

The last layer of the SNN consists of $N_\text{out}$ neurons for each action dimension, respectively. The decoder layer is made up of non-spiking integrate neurons connected to the last layer of SNN:
\begin{equation}
    V_t^{out}=V_{t-1}^{out} + W^{out}\cdot S_t^L + b^{out},
\end{equation}
where $W^{out}$ and $b^{out}$ are weights and biases. The final output action is determined by the membrane potential in the last time step $a=V_T^{out}$. The detailed forward propagation of the spiking actor network is shown in Algo.~\ref{algo:SAN}.
\begin{algorithm}
    \caption{Forward propagation of spiking actor network}
    \label{algo:SAN}
    \begin{algorithmic}[1]
        \STATE \textbf{Input:} $M_s$-dimensional observation $s$
        \STATE Compute the stimulation of input populations: $$ A_E = \exp\left[-\frac{1}{2}\frac{(s-\mu)^2}{\sigma^2}\right]$$

         \FOR{t=1,\dots,T}
         \STATE Compute the output spikes of the population encoder: $$V_t^{in} = V_{t-1}^{in}- S_{t-1}^{in} + A_E$$$$
S_t^{in} = \Theta(V_t^{in} - V_E)$$
         \FOR{l=1,\dots,L}
         \STATE Update neurons in layer $l$ at timestep $t$
         \ENDFOR
         \STATE Update the decoder neurons' membrane potential: $$V_t^{out}=V_{t-1}^{out} + W^{out}\cdot S_t^L + b^{out}$$
         \ENDFOR
        \STATE \textbf{Output:} $M_a$-dimensional action $a=V_T^{out}$
    \end{algorithmic}
\end{algorithm}
\subsubsection{Back propagation of the SAN}
SAN parameters are trained by the gradient with respect to the output action $\frac{\partial L}{\partial a}$, where $a=V_T^{out}$.

The output decoder can be updated by:
\begin{equation}
    \begin{array}{c}
    \frac{\partial L}{\partial W^{out}}=\frac{\partial L}{\partial a}\cdot\frac{\partial V_T^{out}}{\partial W^{out}}\\
    \frac{\partial L}{\partial b^{out}}=\frac{\partial L}{\partial a}\cdot\frac{\partial V_T^{out}}{\partial b^{out}}
    \end{array}
\end{equation}

Then, the main SNN is trained by STBP with the rectangular surrogate function defined as:
\begin{equation}
    \Theta'(x)=\left\{\begin{matrix} 
  \frac{1}{2\omega} , &-\omega\le x\le\omega  \\  
  0,& \text{else} 
\end{matrix}\right. ,
\label{eq:SG}
\end{equation}
where $\omega$ is the window size. 

Next, the gradient of the encoder stimulation $A_E$ is written in Eq.\ref{eq:grad_AE}. Note that $\frac{\partial S_t^{in}}{\partial A_E }$ is manually set to $1$ to simplify the gradient computation.
\begin{equation}
    \frac{\partial L}{\partial A_E}=\sum_{t=1}^{T}  \frac{\partial L}{\partial S_t^{in}}\cdot \frac{\partial S_t^{in}}{\partial A_E }=\sum_{t=1}^{T}  \frac{\partial L}{\partial S_t^{in}}
    \label{eq:grad_AE}
\end{equation}
Finally, the trainable parameters $\mu$ and $\sigma$ in the encoder can be updated by:
\begin{equation}
    \begin{array}{c}
\frac{\partial L}{\partial \mu }=\frac{\partial L}{\partial A_E}\cdot \frac{\partial A_E}{\partial \mu} = \frac{\partial L}{\partial A_E} \cdot  \frac{s-\mu}{\sigma^2} A_E\\
\frac{\partial L}{\partial \sigma }=\frac{\partial L}{\partial A_E}\cdot \frac{\partial A_E}{\partial \sigma}=\frac{\partial L}{\partial A_E} \cdot \frac{(s-\mu)^2}{\sigma^3} A_E
    \end{array}
\end{equation}

\subsection{Other Spiking Neuron Models}
Section \ref{sec:SNN} already shows the LIF neuron model, this section will show two other spiking neuron models conducted in the experiments.
\subsubsection{Current-Based LIF neuron model}
In the current-based LIF (CLIF) neurons proposed in \cite{popSAN}, the input current in Eq.\ref{eq:current} is redefined as:
\begin{equation}
    I_t^l=\alpha I_{t-1}^l + W^lS_t^{l-1}+b^l,
\end{equation}
where $\alpha$ is the current leakage parameter. While other dynamics of CLIF neurons are the same as those of LIF neurons. 
\subsubsection{Dynamic neuron model}
\cite{MDC_SAN} designed a second-order dynamic neurons (DN) for continuous control. The DN consists of a membrane potential $V$ and a resistance item $U$ to simulate hyperpolarization. Its dynamics is shown as follows:
\begin{equation}
    \frac{d V_t^l}{d t}={V_t^l}^{2}-V_t^l-U_t^l+I_t^l
\end{equation}
\begin{equation}
    \frac{d U_t^l}{d t}=\theta_{v} V_t^l-\theta_{u} U_t^l
\end{equation}
where $\theta_{v}$ and $\theta_{U}$ are the conductivities of $V$ and $U$, respectively. Once firing a spike, the membrane potential $V$ is reset to $V_{\text{reset}}$ and the resistance $U$ is added by $\theta_s$.

With a firs-order Taylor expansion, the iterative DN can be written as:
\begin{equation}
    \begin{array}{l}
 {C}_{t}^{l}=\alpha  \cdot  {C}_{t-1}^{l}+ {W}^{l}  {S}_{t}^{l-1}+ {b}^{l} ; \\
 {V}_{t}^{l}= \left(1- {S}_{t-1}^{l}\right) \cdot{V}_{t-1}^{l} + {S}_{t-1}^{l} \cdot V_{\text{reset}} ; \\
 {U}_{t}^{l}= {U}_{t-1}^{l}+ {S}_{t-1}^{l} \cdot \theta_{u} ; \\
 {V}_{\text {delta }}= {V}_{t}^{l^{2}}- {V}_{t}^{l}- {U}_{t}^{l}+ {C}_{t}^{l} ; \\
 {U}_{\text {delta }}=\theta_{v} \cdot  {V}_{t}^{l}-\theta_{u} \cdot  {U}_{t}^{l} ; \\
 {V}_{t}^{l}= {V}_{t}^{l}+V_{\text {delta }} ; \\
 {U}_{t}^{l}= {U}_{t}^{l}+U_{\text {delta }} ; \\
 {S}_{t}^{l}=\Theta\left( {V}_{t}^{l}-V_{t h}\right) .
\end{array}
\end{equation}

\subsection{Experiment details}
\label{app:exp_para}
\subsubsection{Compute Resources}
We conduct the experiments on an RTX 3090 GPU and an Intel(R) Xeon(R) Platinum 8362 CPU.
\subsubsection{Spiking Neuron Parameters} The LIF and CLIF neuron parameters are shown in Tab.\ref{tab:LIF_CLIF}, which are the same as those in \cite{popSAN}, except that the LIF neuron has no current leakage parameter. The DN parameters are shown in Tab.\ref{tab:DN}, determined by the pre-learning process proposed in \cite{MDC_SAN}.

\begin{table}[htbp]
  \caption{Parameters of LIF and CLIF \citep{popSAN} neurons}
  \label{tab:LIF_CLIF}
  \centering
  \begin{tabular}{lcc}
  \toprule
 Parameter& LIF&CLIF \citep{popSAN} \\ 
 \midrule
 Membrane leakage parameter $\lambda$&$0.75$&$0.75$ \\
 Threshold voltage $V_{th}$& $0.5$&$0.5$ \\
 Reset voltage $V_{\text{reset}}$& $0$&$0$ \\
 Current leakage parameter $\alpha$& -& $0.5$ \\
 \bottomrule
  \end{tabular}
\end{table}

\begin{table}[htbp]
  \caption{Parameters of the DN \citep{MDC_SAN}}
  \label{tab:DN}
  \centering
  \begin{tabular}{lc}
  \toprule
 Parameter&Value\\ 
 \midrule
 SNN time steps&$5$\\
 Threshold voltage $V_{th}$&$0.5$ \\
 Current leakage parameter $\alpha$&$0.5$ \\
 Conductivity of membrane potential $\theta_v$&$-0.172$\\
 Conductivity of hidden state $\theta_u$&$0.529$\\
 Reset voltage $V_{\text{reset}}$&$0.021$\\
 spike effect to hidden state $\theta_s$&$0.132$\\
 \bottomrule
  \end{tabular}
\end{table}

\subsubsection{Specific Parameters for the Proxy Target Framework} Tab.\ref{tab:proxy} shows hyper-parameters of the proxy target framework for different spiking neurons. To capture the behavior of the SNN, the hidden sizes of the proxy network is set wider than that of its online SNN. Since different spiking neurons exhibit different dynamics and learning speed, hidden sizes\footnote{Since the InvertedDoublePendulum environment is relatively easier, there is no needs for such a wide proxy network. Thus we set the hidden size is $(512,512)$ specifically for that environment for the CLIF neuron.} and learning rate of the proxy network vary across spiking neurons. All other hyper-parameters are kept the same.

\begin{table}[htbp]
  \caption{Hyper-parameters of the proxy network framework with different spiking neurons}
  \label{tab:proxy}
  \centering
  \begin{tabular}{lccc}
  \toprule
 Parameter& LIF& CLIF \citep{popSAN} &DN \citep{MDC_SAN}\\ 
 \midrule
 Proxy network architecture &$(512,512)$& $(800,600)$&$(512,512)$\\
 Proxy network activation& ReLU& ReLU&ReLU\\
  Proxy network learning rate& $1\cdot 10^{-3}$& $3\cdot 10^{-3}$&$3\cdot 10^{-3}$\\
 Proxy network optimizer & Adam& Adam&Adam\\
 Proxy update iterations $K$& $3$& $3$&$3$\\
 Proxy update batch size $N$& $256$& $256$&$256$\\
 \bottomrule
  \end{tabular}
\end{table}

\subsubsection{Spiking Actor Network Parameters} All hyper-parameters of the spiking actor network are shown in Tab.\ref{tab:SAN}. This is the same as in a wide range of previous studies \citep{popSAN,MDC_SAN,ILC_SAN}.

\begin{table}[htbp]
  \caption{Hyper-parameters of the spiking actor network}
  \label{tab:SAN}
  \centering
  \begin{tabular}{lc}
  \toprule
 Parameter& Value\\ 
 \midrule
 Encoder population per dimension $N_{in}$&$10$\\
 Encoder threshold $V_E$& $0.999$\\
  Network hidden units& $(256,256)$\\
 Decoder population per dimension $N_{out}$& $10$\\
 Surrogate gradient window size $\omega$& $0.5$\\
 \bottomrule
  \end{tabular}
\end{table}

\subsubsection{RL algorithm parameters} We conduct the experiment based on the TD3 algorithm \citep{TD3}, with hyper-parameters shown in Tab.\ref{tab:TD3}.

\begin{table}[htbp]
  \caption{Hyper-parameters of the implemented TD3 algorithm \citep{TD3}}
  \label{tab:TD3}
  \centering
  \begin{tabular}{lc}
  \toprule
 Parameter& Value \\
 \midrule
 Actor learning rate& $3\cdot10^{-4}$\\
  Actor regularization & None\\
 Critic learning rate& $3\cdot10^{-4}$\\
 Critic regularization & None\\
 Critic architecture &$(256,256)$\\
 Critic activation &ReLU\\
 Optimizer & Adam\\
 Target update rate $\tau$& $5\cdot10^{-3}$\\
 Batch size $N$& $256$\\
 Discount factor $\gamma$& $0.99$\\
 Iterations per time step&$1.0$ \\
 Reward scaling & $1.0$ \\
 Gradient clipping & None \\
 Replay buffer size& $10^{6}$ \\
 Exploration noise $\mathcal{N}(0,\sigma)$& $\mathcal{N}(0,0.1)$\\
 Actor update interval $d$& $2$\\
 Target policy noise $\mathcal{N}(0,\tilde{\sigma})$& $\mathcal{N}(0,0.2)$\\
 Target policy noise clip $c$& $0.5$\\
 \bottomrule
  \end{tabular}
\end{table}

\subsubsection{Experiment environments}

% \begin{figure}
%   \centering
% \includegraphics[width=1\linewidth]{PDFs/envs.pdf}
%   \caption{Several continuous control tasks of the MuJoCo environments on OpenAI Gymnasium. (a) InvertedDoublePendulum-v4, (b) Ant-v4, (c) HalfCheetah-v4, (d) Hopper-v4, (e) Walker2d-v4.}
%   \label{fig:envs}
% \end{figure}

Fig.~\ref{fig:envs} shows various MuJoCo environments \citep{mujoco1,mujoco2} on OpenAI Gymnasium benchmarks \citep{gym,gymnasium}, including InvertedDoublePendulum (IDP) \citep{IDP}, Ant \citep{GAE}, HalfCheetah \citep{halfcheetah}, Hopper \citep{hopper} and Walker2d. All environment setups used the default configurations without modifications.

It is worth noting that the state vector ranges from $-\infty$ to $\infty$, it is normalized to $(-1,1)$ by a tanh function. In addition, since the action has the minimum and maximum limits, the output of actor network is normalized to $(-1,1)$ by a tanh function and then linearly scaled to $(\text{Min action},\text{Max action})$.
\subsection{Pseudo codes for the proposed proxy target framework in conjunction with TD3}
\label{app:RL_framework}
We present the detailed pseudocode of the general proxy target framework in Algo.\ref{algo:general}, in Section \ref{sec:overall_algo}. Specifically, Algo.\ref{algo:TD3} shows how to implement the proxy target framework in the TD3 algorithm \citep{TD3}. It is worth noting that the original TD3 algorithm updates the target actor with delay. However, in our framework, the proxy actor is updated without delay because of its inherently slow update pace.

\begin{algorithm}[H]
    \caption{Proxy target framework with TD3}
    \label{algo:TD3}
    \begin{algorithmic}[1]
        \STATE Initialize SNN actor network $\pi_\phi^{\text{SNN}}(s)$, ANN critic networks $Q_{\theta_1}^{\text{ANN}}(s,a)$, $Q_{\theta_2}^{\text{ANN}}(s,a)$ with weights $\phi$, $\theta_1$ and $\theta_2$
        \STATE Initialize proxy actor $\pi_{\phi'}^{\text{Proxy}}(s)$ and ANN target critics $Q_{\theta_1'}^{\text{ANN}}(s,a)$, $Q_{\theta_2'}^{\text{ANN}}(s,a)$ with weights $\phi'$, $\theta_1'$ and $\theta_2'$
        \STATE Initialize replay buffer $\mathcal{D}$
        \FOR{each iteration}
        \STATE Execute action $a=\pi_\phi^{\text{SNN}}(s)+\epsilon$, $\epsilon\sim\mathcal{N}(0, \sigma)$ and observe reward $r$ and next state $s'$
        \STATE Store the transition $(s, a, r, s')$ in $\mathcal{D}$
        \FOR{$k=1$ to $K$}
            \STATE Sample a minibatch of $N$ transitions $(s_i, a_i, r_i, s'_i)$ from $\mathcal{D}$
            \STATE Update proxy actor network parameters $\phi'$ by minimizing the loss:$$L_{target}=\frac{1}{N} \sum_i \left\| \pi_{\phi'}^{\text{Proxy}}(s_i) - \pi_\phi^{\text{SNN}}(s_i) \right\|_2^2 $$
        \ENDFOR
        
        \STATE $y_i=r_i+\gamma\min_{j=1,2} Q_{\theta_j'}^{\text{ANN}}(s_i,\tilde{a}_i)$, $\tilde{a}_i=\pi_{\phi'}^{\text{Proxy}}(s_i)+\epsilon$,  $\epsilon\sim\text{clip}\left(\mathcal{N}(0,\tilde{\sigma}), -c,c\right)$  
        \STATE Update ANN critics by minimizing the critic loss: $L_{critic}=\frac{1}{N}\sum_i\left(Q_{\theta_j}^{\text{ANN}}(s_i,a_i)-y_i\right)^2$

        \IF{$t$ mod $d$}
            \STATE Update SNN actor by maximizing the objective function: $J=\frac{1}{N}\sum_iQ_{\theta_1}^{\text{ANN}}\left(s_i,\pi_\phi^{\text{SNN}}(s_i)\right)$
            \STATE Update ANN critic target parameters explicitly by the Polyak function: $\theta_j'\gets\tau\theta_j + (1-\tau)\theta_j'$
  
        \ENDIF
        \ENDFOR
    \end{algorithmic}
\end{algorithm}

\subsection{Additional experiments results}
\subsubsection{Additional results in terms of performance}
\begin{figure}[H]
  \centering
\includegraphics[width=1\linewidth]{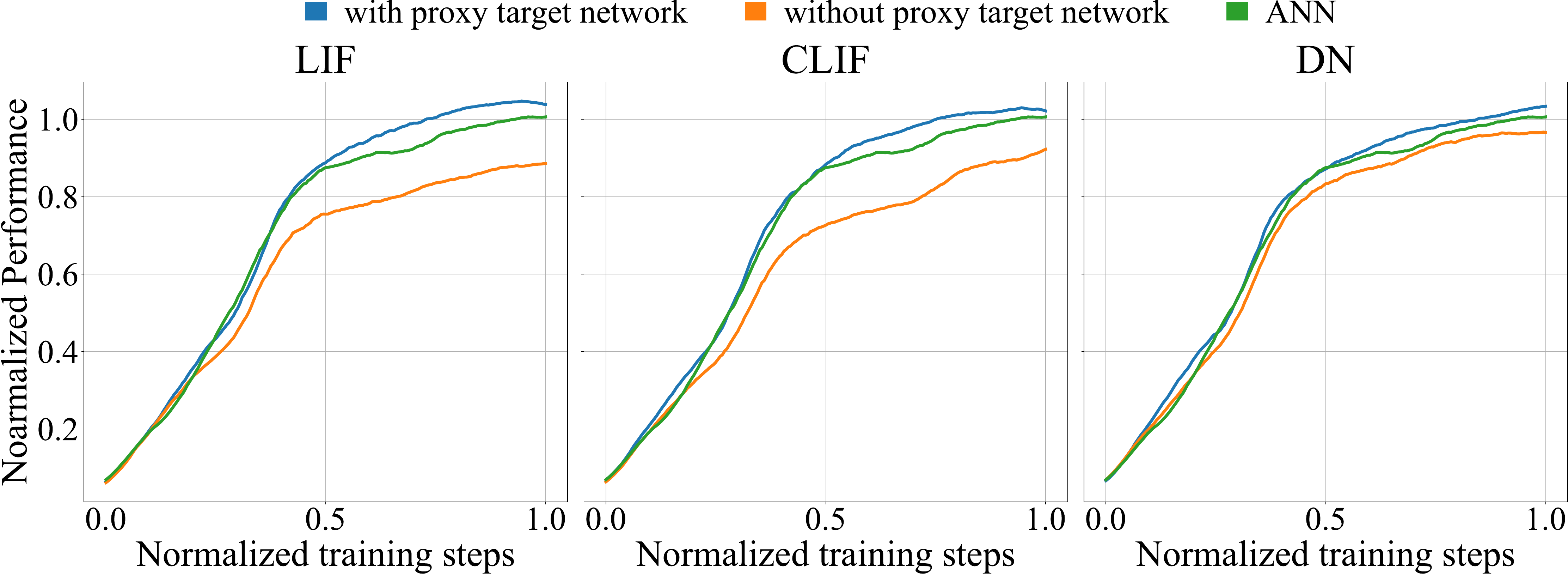}
  \caption{Normalized learning curves of the proposed proxy target framework with different spiking neurons across all environments. The performance and training steps are normalized linearly based on ANN performance. Curves are uniformly smoothed for visual clarity.}
  \label{fig:cmp_overall}
\end{figure}

In the main text, we show that our proxy target framework can increase performance for various spiking neurons. Fig.~\ref{fig:cmp_overall} shows the normalized learning curves of our proxy target framework for different spiking neurons. In addition, Tab.~\ref{tab:cmp_snn:LIF}, Tab.~\ref{tab:cmp_snn:CLIF}, and Tab~.\ref{tab:cmp_snn:DN} show the maximum average returns and the average performance gains of the proxy network against vanilla SNN with LIF neuron, CLIF neuron, and dynamic neuron, respectively. 

\begin{table}[htbp]
  \footnotesize
  \caption{Max average returns over $5$ random seeds with LIF neurons.}
  \label{tab:cmp_snn:LIF}
  \centering
  \begin{tabular}{lcccccc}
    \toprule
    Method&IDP&Ant&HalfCheetah&Hopper&Walker2d& APG\\ \midrule
      Vanilla LIF& $9347 \pm 1$& $4294 \pm 1170$&$9404 \pm 625$& $\mathbf{3520 \pm 94}$& $1862 \pm 1450$& \multirow{2}{*}{$\mathbf{32.15\%}$}\\
         PT-LIF&$\mathbf{9348 \pm 1}$&$\mathbf{5383 \pm 250}$&$\mathbf{10103 \pm 607}$&$3385 \pm 157$&$\mathbf{4314 \pm 423}$&\\
    % \midrule
    \bottomrule
  \end{tabular}
 \end{table}

 \begin{table}[htbp]
  \footnotesize
  \caption{Max average returns over $5$ random seeds with CLIF neurons.}
  \label{tab:cmp_snn:CLIF}
  \centering
  \begin{tabular}{lcccccc}
    \toprule
    Method&IDP&Ant&HalfCheetah&Hopper&Walker2d& APG\\ \midrule
      Vanilla CLIF& $\mathbf{9351 \pm 1}$& $4590 \pm 1006$&$9594 \pm 689$& $2772 \pm 1263$& $3307 \pm 1514$& \multirow{2}{*}{$\mathbf{15.03\%}$}\\
         PT-CLIF&$\mathbf{9351 \pm 1}$&$\mathbf{5014 \pm 1074}$&$\mathbf{9663 \pm 426}$&$\mathbf{3526 \pm 112}$&$\mathbf{4564 \pm 555}$&\\
    \bottomrule
  \end{tabular}
 \end{table}

 \begin{table}[H]
  \footnotesize
  \caption{Max average returns over $5$ random seeds with dynamic neurons.}
  \label{tab:cmp_snn:DN}
  \centering
  \begin{tabular}{lcccccc}
    \toprule
    Method&IDP&Ant&HalfCheetah&Hopper&Walker2d& APG\\ \midrule
      Vanilla DN& $\mathbf{9350 \pm 1}$& $4800 \pm 994$&$9147 \pm 231$& $3446 \pm 131$& $3964 \pm 1353$& \multirow{2}{*}{$\mathbf{4.87\%}$}\\
         PT-DN&$\mathbf{9350 \pm 1}$&$\mathbf{5400 \pm 277}$&$\mathbf{9347 \pm 666}$&$\mathbf{3507 \pm 144}$&$\mathbf{4277 \pm 650}$&\\
    \bottomrule
  \end{tabular}
 \end{table}

\subsubsection{Additional results in ANN}
\begin{figure}[H]
  \centering
\includegraphics[width=1\linewidth]{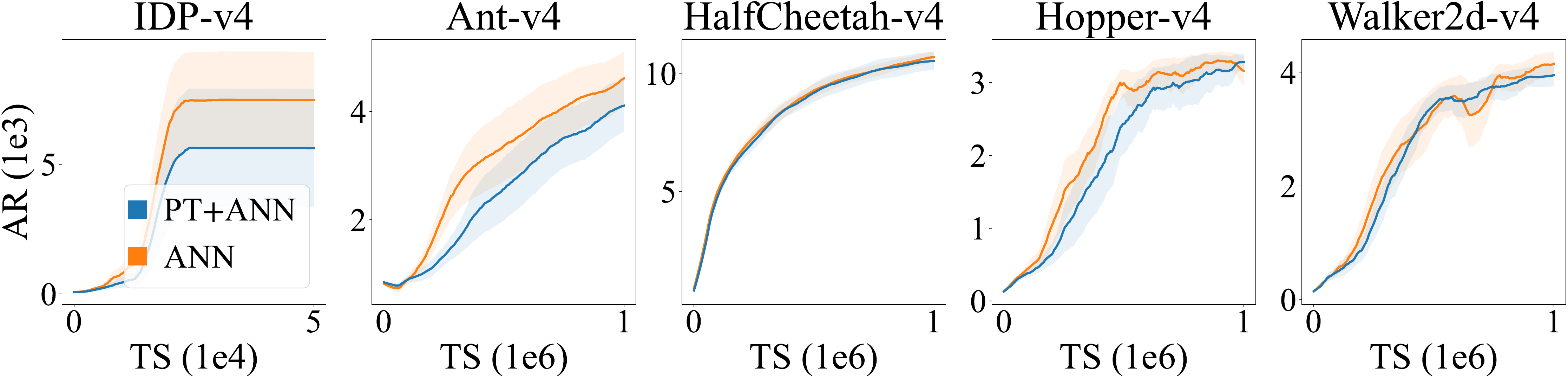}
  \caption{Learning curves of utilizing the proxy target framework in ANN. The PT represents the proxy target framework, AR denotes average returns, and TS is training steps. The shaded region represents half a standard deviation over 5 different seeds. Curves are uniformly smoothed for visual clarity.}
  \label{fig:detailed_ann_curves}
\end{figure}

We show the normalized learning curves of the proxy target framework with ANN in Fig.~\ref{fig:var_ablation}(b). Here, we show the detailed learning curves and maximum average returns of $5$ environments in Fig.~\ref{fig:detailed_ann_curves} and Tab.\ref{tab:cmp_snn:ann}, respectively.

 \begin{table}[H]
  \footnotesize
  \caption{Max average returns over $5$ random seeds with ANN (TD3).}
  \label{tab:cmp_snn:ann}
  \centering
  \resizebox{1.0\textwidth}{!}{
  \begin{tabular}{lcccccc}
    \toprule
    Method&IDP&Ant&HalfCheetah&Hopper&Walker2d& APG\\ \midrule
      ANN& $\mathbf{7503 \pm 3713}$& $\mathbf{4770 \pm 1014}$&$\mathbf{10857 \pm 475}$& $3410 \pm 164$& $\mathbf{4340 \pm 383}$& \multirow{2}{*}{$\mathbf{-8.38\%}$}\\
         PN-ANN&$5653 \pm 4540$&$4234 \pm 998$&$10708 \pm 773$&$\mathbf{3435 \pm 145}$&$4106 \pm 366$&\\
    % \midrule
    \bottomrule
  \end{tabular}
  }
 \end{table}

\end{document}